# Design Challenges for Robots in Industrial Applications

**Nesreen Mufid, EiZ Engineering, Amman, Jordan. Email: nesreen@eiz-eng.com.jo**

**1. Introduction**

      Nowadays, electric robots play big role in many fields as they can replace humans and/or decrease the amount of load on humans. There are several types of robots that are present in the daily life, some of them are fully controlled by humans while others are programmed to be self-controlled. In addition there are self-control robots with partial human control. Robots [1] can be classified into three major kinds: industry robots, autonomous robots and mobile robots.

      Industry robots are used in industries and factories to perform mankind tasks in the easier and faster way which will help in developing products. Typically industrial robots perform difficult and dangerous tasks, as they lift heavy objects, handle chemicals, paint and assembly work and so on. They are working all the time hour after hour, day by day with the same precision and they don't get tired which means that they don't make errors due to fatigue. Indeed, they are ideally suited to complete repetitive tasks.

      On the other hand, autonomous robots operate independently without the need of human in controlling. They are able to operate and adapt their environment without direct human supervision. Actually, autonomous robots [2] should be able to complete their duties in changing environment and without human control. They self-learn from their daily experience.

      Finally, mobile robots which are robots that are able to move and perform tasks that people cannot operate; For instance, entering dangerous areas, collapsed buildings, nuclear generators (e.g. Fukushima nuclear generator in Japan) and so on. Mobile robots are usually used to observe the environment and discover different places such as The Mars Exploration Program (MEP) [3] which is led by NASA. MEP uses spacecraft, landers and rovers to discover the possibilities of life in Mars. The Mars rover, which is an automated motor vehicle, can drive itself upon arrival on the surface of Mars. It has more advantages over the stationary lander like, examining more territory. In fact, Mars rovers improve the knowledge of how to control robotic vehicles remotely.

## 1.1. Motivation

The purpose of this project is to design a mobile robot car that is able to reach the required destination without struggling by any obstacles. This car will be controlled over the cellular system to have the widest coverage possible, to avoid short range communications (e.g. Zigbee [4]), and to overcome the communication drop of WiFi routers in case of disasters [5]. In the other word, it will have two modes; 1) self-control in order to avoid obstacles using sensors and 2) remote controlling over cellular communication using Graphical User Interface (GUI) application, which is accomplished through Android. In addition, camera is installed on the car's top to provide surface monitoring, which allows the car to show the surrounding to the user. In order to make accurate decision, the output of the sensor is also used to find the real distance through calculations.

This project establishes two ways of communications; from the robot to the user showing the video content and status about the scenario. And from the user to the robot indicating the direction, distance and/or offloading demands.

## 1.2. Application

There are many areas where human can't enter due to hazardous and fatal conditions or small dimensions, for instance collapsed buildings, areas after disasters and earthquake, nuclear power plants and so on. For example, the great earthquake that occurred on March 11$^{th}$ 2012 and caused damage to the north part of Japan particularly in Fukushima Daiichi nuclear power plant. The disaster result in disabling the power supply and heat sinks which leads to release of radioactivity in the area surrounding the plant. Such areas and environment are very dangerous to enter by human being, in this case robotic car can be sent in order to search and discover.

## 2. Design criteria

The proposed Model include various features to provide an intelligent and safe environment inside the home for the home members when they are away from it. In order to build such a system, the selection of different electrical devices and components in the system should match our criteria [3]. After a deep studying of the previous projects of smart homes, we have chosen our design criteria based on the challenges discussed previously. These criteria of design are:

1. Time constraint [6]
2. Environmental [7]
3. Economical [8]
4. Sustainability [9]
5. Ethical [10]

### 2.1. Time constraint

This project is restricted to time axis, starting with the literature review research, and the development of the robot body including sensors and shields. Also, the communication system has been started as well as the mobile application. The work is also focused on the communication system, the mobile application and the camera will be installed.

### 2.2. Environmental

Regarding the environmental constrains, the car body allows its movement freely in smooth dry environment. If the environment will change the robotic car must be upgraded in some features (e.g. wheels size and type) to adapt to the new situation.

### 2.3. Economical

From economical point of view, building this robotic car is easily achieved for a single prototype for the project. But if we are looking for a mass production of this robotic car, the cost should be optimized to minimize cost to a certain limit that keeping good quality.

### 2.4. Sustainability

This project can sustain for a long time if it properly used, it can work in all countries that support cellular communication (i.e., Worldwide) and it is able to do task in dry environments.

### 2.5. Ethical

Ethics is a hard rule that must follow. This project contains a communication system and camera those two elements must be properly used without any irritation for others or any spy works. Its extension to combat terrorist groups is possible.

## 3. Design of the robotic car

For such a project the physical dimensions of the car platform in very important because it may affect the robot performance during the mission. The following is a compression *Table(1)* that shows three types of car platform. In this project, Magician Chassis [11] were chosen because it allows a high degree of freedom for microprocessor type in addition to its small size, availability and low price.

**Table 1: Comparison between different car platforms.**

| Name | Arduino Robot | Sparki | Magician Chassis |
|---|---|---|---|
| Physical representation | 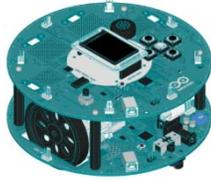 | 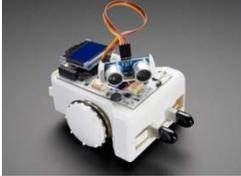 | 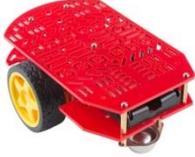 |
| Availability | Not Available | Available | Available |
| Price | $244.95 | $159.95 | $14.95 |
| Size | Radius= 185 mm  Height=85 mm | 90X100 mm | 110X174 mm |
| Processor | ATmega32u4 | Atmega32u4RC | Flexible |

### 3.1. Microcontroller

Microcontroller is a complete computer system on a chip. In Fact, it is the main part of each robotic design. They are used for the operation of embedded systems (computer systems) in robots, motor vehicles, complex medical devices, vending machines and other devices. Each microcontroller includes processor, memory and peripherals. Microcontrollers are programmed to perform simple tasks for other hardware.

On the other hand, Field Programmable Gate Arrays (FPGA) [12], is an integrated circuit that could also be programmed to perform certain task for other hardware. FPGA's contains millions of logic gates that can be configured to do the required tasks. The following *Table(2)* shows comparison between the microcontroller and the FPGA and the reasons for choosing to work with Microcontroller over FPGA. Microcontroller type depends on the requirements of the design and whether it can satisfy them and it will be decided later on.

Table 2: Comparison between Microcontroller and FPGA.

| | FPGA | Microcontroller |
|---|---|---|
| Physical representation | 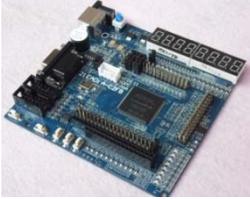 | 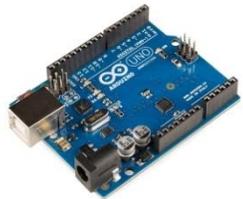 |
| Size | Large | Small |
| Storage | RAM | Flash memory |
| Control Over | Hardware (Designing the hardware – circuit) | Software (chip is already designed) |
| Ability to add/subtract functionality | Flexible | Unable |
| Price | High | Low |
| Pins | Specified by the designer | Dedicated pins for specific features |
| Power consumption | High | Low |

## 4. Conclusions and Challenges

Networks and how they work, IP addresses and how they assigned to each device since we haven't take any network courses. In addition, client-server model was a new idea to go through it, and how to apply it in the project.

The JAVA programming language faced problems to program the mobile application due to this. Also the time constrain play a big role. For this problem more JAVA learning will have the ability to start the application programming with good basics.

The firmware of the WIFI shield has been changed in Arduino 1.0.4 accordingly without updated firmware, so that the shield won't work except with Arduino IDE 1.0.2. On the other hand, there are many complains about updated firmware which says that the shield may not work properly and it's not down-gradable. Thus, the Arduino IDE 1.0.2 has been chosen to be used in this project to reduce the risk of destroying the WiFi shield.